\documentclass{article} % For LaTeX2e
\usepackage{nips13submit_e,times}
\usepackage{url}

\usepackage{epsfig}
\usepackage{graphicx}
\usepackage{amsmath}
\usepackage{amssymb}
\usepackage{enumerate}

\newcommand{\dirnet}[1]{#1.eps}
\newcommand{\dirin}[1]{#1_small.eps}
\newcommand{\dirdet}[1]{#1_small.eps}
\newcommand{\dirres}[1]{#1.eps}
\newcommand{\dirloc}[1]{#1_small.eps}
\newcommand{\dirfigs}[1]{#1.eps}
\newcommand{\fig}[1]{Fig.~\ref{fig:#1}}
\newcommand{\tab}[1]{Table~\ref{tab:#1}}
\newcommand{\secc}[1]{Section~\ref{sec:#1}}
\def\etal{{\textit{et~al.~}}}

\title{OverFeat:\\Integrated Recognition, Localization and Detection\\using Convolutional Networks}

\def\and{%
  \end{tabular}%
  \hskip .01em
  \begin{tabular}[t]{c}}

\author{Pierre Sermanet
  \hskip 1em
  \textbf{David Eigen}\\
  \textbf{Xiang Zhang}
  \hskip 1em
  \textbf{Michael Mathieu}
  \hskip 1em
  \textbf{Rob Fergus}
  \hskip 1em
  \textbf{Yann LeCun}\\
Courant Institute of Mathematical Sciences, New York University\\
719 Broadway, 12th Floor, New York, NY 10003\\
\texttt{{sermanet,deigen,xiang,mathieu,fergus,yann}@cs.nyu.edu}
}

\nipsfinalcopy % Uncomment for camera-ready version

\begin{document}

\maketitle

\begin{abstract}
We present an integrated framework for using Convolutional Networks
for classification, localization and detection. We show how a
multiscale and sliding window approach can be efficiently
implemented within a ConvNet.
We also introduce a novel deep learning approach to localization
by learning to predict object boundaries. Bounding boxes are then
accumulated rather than suppressed in order to increase detection confidence.
We show that different tasks can be learned simultaneously using a single
shared network.
This integrated framework is the winner of the localization task of the
ImageNet Large Scale Visual Recognition Challenge 2013 (ILSVRC2013)
and obtained very competitive results for the detection and classifications
tasks. In post-competition work, we establish a new state of the art
for the detection task.
Finally, we release a feature extractor from our best model called OverFeat.
\end{abstract}

\section{Introduction}

Recognizing the category of the dominant object in an image is a tasks
to which Convolutional Networks (ConvNets)~\cite{lecun-98} have been
applied for many years, whether the objects were handwritten
characters~\cite{lecun-90c}, house numbers~\cite{sermanet-icpr-12}, textureless
toys~\cite{lecun-04}, traffic
signs~\cite{Ciresan12,sermanet-ijcnn-11}, objects from the
Caltech-101 dataset~\cite{jarrett-iccv-09}, or objects from the
1000-category ImageNet dataset~\cite{Kriz12}. The accuracy
of ConvNets on small datasets such as Caltech-101, while decent, has
not been record-breaking. However, the advent of larger datasets has
enabled ConvNets to significantly advance the state of the art on
datasets such as the 1000-category ImageNet \cite{imagenet}.

The main advantage of ConvNets for many such tasks is that the entire
system is trained {\em end to end}, from raw pixels to ultimate
categories, thereby alleviating the requirement to manually design a
suitable feature extractor. The main disadvantage is their ravenous
appetite for labeled training samples.

The main point of this paper is to show that training a convolutional network
to simultaneously classify, locate and detect objects in images can boost
the classification accuracy and the detection and localization
accuracy of all tasks. The paper proposes a new integrated approach to object
detection, recognition, and localization with a single
ConvNet. We also introduce a novel method for localization and detection
by accumulating predicted bounding boxes.
We suggest that by combining many localization predictions,
detection can be performed without training on background samples
and that it is possible to avoid the time-consuming and complicated
bootstrapping training passes. Not training on background also
lets the network focus solely on positive classes for higher accuracy.
Experiments are conducted on the ImageNet ILSVRC 2012 and 2013 datasets
and establish state of the art results on the ILSVRC 2013 localization
and detection tasks.

While images from the ImageNet classification dataset are largely chosen to contain a
roughly-centered object that fills much of the image, objects of
interest sometimes vary significantly in size and position within the
image. The first idea in addressing this is to apply a ConvNet at multiple locations in the
image, in a sliding window fashion, and over multiple scales. Even with
this, however, many viewing windows may contain a
perfectly identifiable portion of the object (say, the head of a dog),
but not the entire object, nor even the center of the object. This leads
to decent classification but poor localization and detection.
Thus, the second idea is to train the system to not only produce a
distribution over categories for each window, but also to produce a
prediction of the location and size of the bounding box containing the
object relative to the window. The third idea is to
accumulate the evidence for each category at each location and size.

Many authors have proposed to use ConvNets for detection and
localization with a sliding window over multiple scales, going back to
the early 1990's for multi-character strings~\cite{matan-92},
faces~\cite{vaillant-monrocq-lecun-94}, and
hands~\cite{nowlan-platt-95}. More recently, ConvNets have been shown
to yield state of the art performance on text detection in natural
images~\cite{delakis-garcia-08}, face
detection~\cite{garcia-delakis-04,osadchy-07} and pedestrian
detection~\cite{sermanet-cvpr-13}.

Several authors have also proposed to train ConvNets to directly
predict the instantiation parameters of the objects to be located,
such as the position relative to the viewing window, or the pose of
the object. For example Osadchy \etal~\cite{osadchy-07} describe a
ConvNet for simultaneous face detection and pose estimation. Faces are
represented by a 3D manifold in the nine-dimensional output
space. Positions on the manifold indicate the pose (pitch, yaw, and
roll). When the training image is a face, the network is trained to
produce a point on the manifold at the location of the known pose. If
the image is not a face, the output is pushed away from the manifold.
At test time, the distance to the manifold indicate whether the image
contains a face, and the position of the closest point on the manifold
indicates pose.
Taylor \etal~\cite{taylor-nips-11,taylor-cvpr-11}
use a ConvNet to estimate the location of body parts (hands, head,
etc) so as to derive the human body pose. They use a metric learning
criterion to train the network to produce points on a body pose
manifold.
Hinton et al. have also proposed to train networks to
compute explicit instantiation parameters of features as part of a
recognition process~\cite{hinton2011transforming}.

Other authors have proposed to perform object localization via
ConvNet-based segmentation. The simplest approach consists in training
the ConvNet to classify the central pixel (or voxel for volumetric
images) of its viewing window as a boundary between regions or
not~\cite{jain-iccv-07}. But when the regions must be categorized, it
is preferable to perform {\em semantic segmentation}. The main idea
is to train the ConvNet to classify the central pixel of the viewing
window with the category of the object it belongs to, using the window
as context for the decision. Applications range from biological image
analysis~\cite{ning-05}, to obstacle tagging for mobile
robots~\cite{hadsell-jfr-09} to tagging of
photos~\cite{farabet-pami-13}. The advantage of this approach is that
the bounding contours need not be rectangles, and the regions need not
be well-circumscribed objects. The disadvantage is that it requires
dense pixel-level labels for training.
This segmentation pre-processing or object proposal step has recently gained popularity
in traditional computer vision to reduce the search space of
position, scale and aspect ratio for detection
~\cite{manen2013prime,cpmc-release1,endres2010category,UijlingsIJCV2013}.
Hence an expensive
classification method can be applied at the optimal location in the search
space, thus increasing recognition accuracy.
Additionally, ~\cite{UijlingsIJCV2013,carreira2012object} suggest
that these methods improve accuracy by drastically reducing
unlikely object regions, hence reducing potential false positives.
Our dense sliding window method, however, is able to outperform
object proposal methods on the ILSVRC13 detection dataset.

Krizhevsky \etal \cite{Kriz12} recently demonstrated impressive
classification performance using a large ConvNet. The authors also
entered the ImageNet 2012 competition, winning both the
classification and localization challenges. Although they demonstrated
an impressive localization performance, there has been no published
work describing how their approach. Our paper is thus the first to
provide a clear explanation how ConvNets can be used for localization
and detection for ImageNet data.

In this paper we use the terms localization and detection in a way
that is consistent with their use in the ImageNet 2013 competition,
namely that the only difference is the evaluation criterion used and both
involve predicting the bounding box for each object in the image. 

%used a separate
%regression network, taking as input classification features, to
%predict the bounding-box extent of an object. Our approach is similar,
%but with important implementation differences. 
% % YLC->Rob: this section might need to be eliminated or appear elsewhere.
% \subsection{Convolutional Nets}

% ConvNets are multi-stage feed-forward architectures in which each
% stage generally contains a linear filter bank with trainable
% coefficients, followed by a point-wise non-linearity (often a
% half-wave rectification), and optionally followed by a spatial pooling
% and subsampling operation.  The main advantage of ConvNets for many
% such tasks is that the entire system is trained {\em end to end}, from
% raw pixels to ultimate categories, thereby alleviating the requirement
% to manually design a suitable feature extractor. The features learned
% by the successive stages are increasingly global, abstract, and
% invariant.

% Training often uses a form of stochastic gradient descent to minimize
% a supervised objective function, the gradient of the objective with
% respect to the filter coefficient being computed with the
% back-propagation method~\cite{}.

% Many of these recent advances have been enabled by the availability of
% large labeled datasets and of fast numerical engines, particularly
% GPU-based hardware.

\begin{figure}[hbt!]
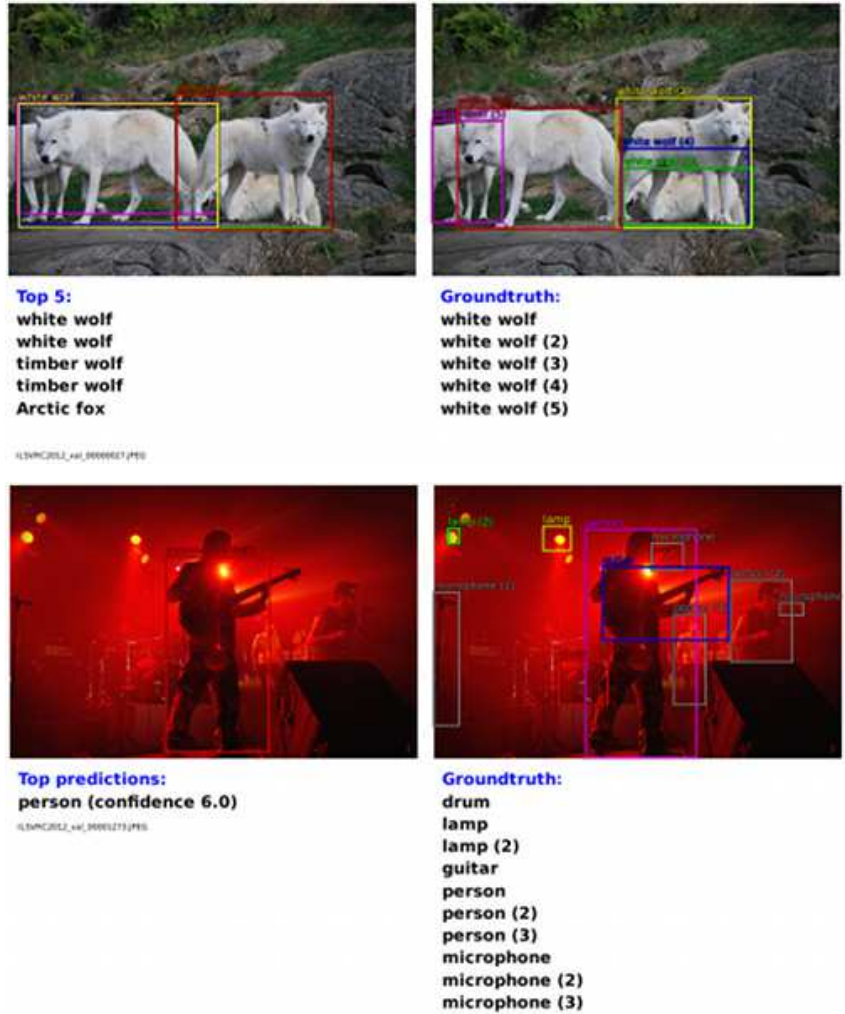

\vspace{-3mm}
\begin{center}
\includegraphics[width=.8\linewidth]{\dirin{loc2}} \\
\vspace{2mm}
\includegraphics[width=.8\linewidth]{\dirin{det13}} \\
\end{center}
\vspace*{-0.3cm}
\caption{\textbf{Localization (top) and detection tasks (bottom).}
The left images contains our predictions (ordered by decreasing confidence)
while the right images show the groundtruth labels. The detection image (bottom)
illustrates the higher difficulty of the detection dataset, which can contain many
small objects while the classification and localization images typically
contain a single large object.}
\label{fig:tasks}
%\vspace*{-0.6cm}
\end{figure}

\section{Vision Tasks}

In this paper, we explore three
computer vision tasks in increasing order of difficulty:
(\emph{i}) classification, (\emph{ii}) localization, and (\emph{iii}) detection.
Each task is a sub-task of the next. While all tasks are adressed using
a single framework and a shared feature learning base, we will describe them separately
in the following sections.

Throughout the paper, we report results on the 2013
ImageNet Large Scale Visual Recognition Challenge (ILSVRC2013).
In the classification task of this challenge, each image is assigned a single label corresponding
to the main object in the image. Five guesses are allowed to find the correct answer
(this is because images can also contain multiple unlabeled objects).
The localization task is similar in that 5 guesses are allowed per image,
but in addition, a bounding box for the predicted object must be returned with each guess.
To be considered correct, the predicted box must
match the groundtruth by at least 50\% (using the PASCAL criterion of union over intersection),
as well as be labeled with the correct class (i.e. each prediction is a label and bounding box that are associated together).
The detection task differs from localization in that there can be any number of objects
in each image (including zero), and false positives are penalized
by the mean average precision (mAP) measure.
The localization task is a convenient intermediate step between classification and
detection, and allows us to evaluate our localization method independently of
challenges specific to detection (such as learning a background class).
In \fig{tasks}, we show examples of images with our localization/detection predictions
as well as corresponding groundtruth.  Note that classification and localization
share the same dataset, while detection also has additional data where objects
can be smaller. The detection data also contain a set of images where certain
objects are absent. This can be used for bootstrapping, but we have not made use of it
in this work.

%%%%%%%%%%%%%%%%%%%%%%%%%%%%%%%%%%%%%%%%%%%%%%%%%%%%%%%%%%%%%%%%%%%%%%%%%%%%%%%%%%%%%%%

\section{Classification}
\label{sec:classify}

Our classification architecture is similar to the best ILSVRC12 architecture
by Krizhevsky \etal \cite{Kriz12}. However, we improve on the network design
and the inference step. Because of time constraints,
some of the training features in Krizhevsky's
model were not explored, and so we expect our results can be improved even further.
These are discussed in the future work section~\ref{sec:discussion}

\begin{figure}[h!]
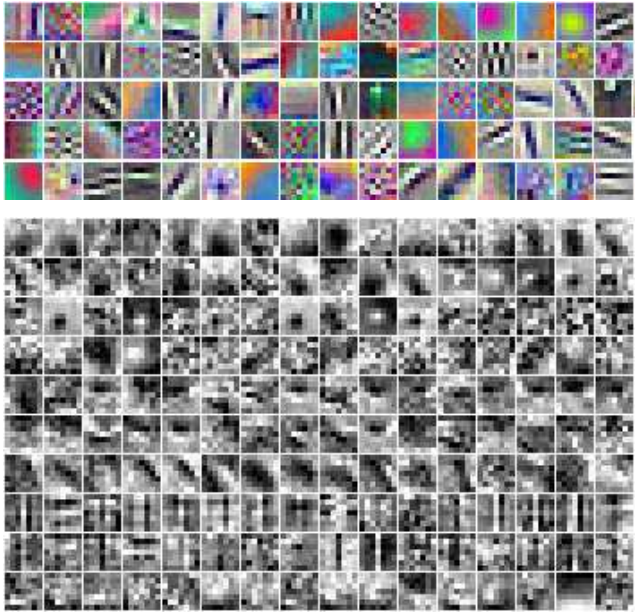

\vspace{-3mm}
\begin{center}
\includegraphics[width=3.3in]{\dirfigs{layer1}} \\
\vspace{2mm}
\includegraphics[width=3.3in]{\dirfigs{layer2}}
\end{center}
\vspace*{-0.3cm}
\caption{\textbf{Layer 1 (top) and layer 2 filters (bottom).}}
\label{fig:features}
%\vspace*{-0.6cm}
\end{figure}

\subsection{Model Design and Training}
\label{sec:model_training}
We train the network on the ImageNet 2012 training set (1.2 million
images and $C=1000$ classes) \cite{imagenet}. Our model uses the same fixed input size
approach proposed by Krizhevsky \etal \cite{Kriz12} during training but turns to
multi-scale for classification as described in the next section. Each image is
downsampled so that the smallest dimension is 256 pixels. We then
extract 5 random crops (and their horizontal flips) of size 221x221
pixels and present these to the network in mini-batches of size
128. The weights in the network are initialized randomly with
$(\mu,\sigma)=(0,1 \times 10^{-2})$. They are then updated by stochastic gradient
descent, accompanied by momentum term of $0.6$ and an $\ell_2$ weight decay of
$1 \times 10^{-5}$. The learning rate is initially $5 \times 10^{-2}$ and is
successively decreased by a factor of $0.5$ after $(30,50,60,70,80)$ epochs.
DropOut \cite{Hinton12} with a rate of $0.5$ is employed on the fully connected layers (6th
and 7th) in the classifier.

\begin{table}[h!]
%\small
\scriptsize
%\vspace*{-2mm}
\begin{center}
\begin{tabular}{|l||c|c|c|c|c||c|c||c|}
  \hline  
         &     &   &   &   &   &   &   & Output \\ 
  Layer  &   1 & 2 & 3 & 4 & 5 & 6 & 7 & 8 \\ \hline
  \hline
  Stage & conv + max & conv + max & conv & conv & conv + max & full & full & full \\ \hline
  \# channels &   96 & 256 & 512 & 1024 & 1024 & 3072 & 4096  & 1000 \\ \hline
  Filter size &  11x11 & 5x5 & 3x3 & 3x3 & 3x3 & - & - & - \\ \hline
  Conv. stride &  4x4 & 1x1 & 1x1 & 1x1 & 1x1 & - & - & -  \\ \hline
  Pooling size &  2x2 & 2x2 & - & - & 2x2 & - & - & -  \\ \hline
  Pooling stride& 2x2 & 2x2 & - & - & 2x2 & - & - & -  \\ \hline
  Zero-Padding size &  - & - & 1x1x1x1 & 1x1x1x1 & 1x1x1x1 & - & - & -  \\ \hline
  Spatial input size & 231x231 & 24x24 & 12x12 & 12x12 & 12x12 &  6x6 & 1x1 & 1x1    \\ \hline
 
%  Constrast norm. & - & \checkmark & \checkmark & \xmark
  \hline
\end{tabular}
\vspace*{2mm}
\caption{\textbf{Architecture specifics for {\em fast} model.} The spatial
  size of the feature maps depends on the input image size, which
  varies during our inference step (see \tab{multiscale} in the Appendix).
  Here we show training spatial sizes. Layer 5 is
  the top convolutional layer. Subsequent layers are fully
  connected, and applied in sliding window fashion at test time.
  The fully-connected layers can also be seen as 1x1
  convolutions in a spatial setting.  Similar sizes for {\em accurate} model
  can be found in the Appendix.}
\label{tab:arch}
%\vspace*{-4mm}
\end{center}
\end{table}

We detail the architecture sizes in tables~\ref{tab:arch} and~\ref{tab:arch2}.
Note that during training,
we treat this architecture as non-spatial (output maps of size 1x1),
as opposed to the inference step, which produces spatial outputs. Layers 1-5 are similar to
Krizhevsky \etal \cite{Kriz12}, using rectification (``{\em relu}'') non-linearities and
max pooling, but with the following differences: (i) no
contrast normalization is used; (ii) pooling regions are
non-overlapping and (iii) our model has larger 1st and 2nd layer
feature maps, thanks to a smaller stride (2 instead of 4). A larger stride
is beneficial for speed but will hurt accuracy.

In \fig{features}, we show the filter coefficients from the first two
convolutional layers. The first layer filters capture orientated
edges, patterns and blobs. In the second layer, the filters have a
variety of forms, some diffuse, others with strong line structures or
oriented edges.

\subsection{Feature Extractor}

Along with this paper, we release a feature extractor
named ``OverFeat''~\footnote{\urlstyle{same}\url{http://cilvr.nyu.edu/doku.php?id=software:overfeat:start}}
in order to provide powerful features for computer vision research.
Two models are provided, a {\em fast} and {\em accurate} one. Each architecture is
described in tables~\ref{tab:arch} and~\ref{tab:arch2}. We also compare
their sizes in \tab{connections} in terms of parameters and connections.
The {\em accurate} model is more accurate than the {\em fast} one (14.18\% classification error
as opposed to 16.39\% in \tab{clsval}), however it requires nearly twice as many connections.
Using a committee of 7 {\em accurate} models reaches 13.6\% classification error as shown in \fig{cls}.

\subsection{Multi-Scale Classification}
\label{sec:ms_class}

In \cite{Kriz12}, multi-view voting is used to boost performance:
a fixed set of 10 views (4 corners and center, with horizontal flip)
is averaged. However, this approach can ignore many regions of the image,
and is computationally redundant when views overlap.
Additionally, it is only applied at a single scale, which may not be the scale
at which the ConvNet will respond with optimal confidence.

Instead, we explore the entire image by densely running the network
at each location and at multiple scales.
While the sliding window approach may be computationally prohibitive for certain
types of model, it is inherently efficient in the case of ConvNets
(see section~\ref{sec:convnet}). This approach yields significantly more views for voting,
which increases robustness while remaining efficient.
The result of convolving a ConvNet on an image of arbitrary size is a spatial map
of $C$-dimensional vectors at each scale.

However, the total subsampling ratio in the network described above is 2x3x2x3, or 36.
Hence when applied densely, this architecture can only produce a classification vector
every 36 pixels in the input dimension along each axis. This coarse distribution
of outputs decreases performance compared to the 10-view scheme because
the network windows are not well aligned with the objects in the images.
The better aligned the network window and the object, the strongest the confidence
of the network response. To circumvent this problem, we take an approach
similar to that introduced by Giusti \etal \cite{giusti-ijcnn-13}, and apply the last subsampling
operation at every offset.  This removes the loss of resolution from this layer,
yielding a total subsampling ratio of x12 instead of x36.

\begin{figure}[h!]
\vspace{-3mm}
\begin{center}
\includegraphics[width=3.3in]{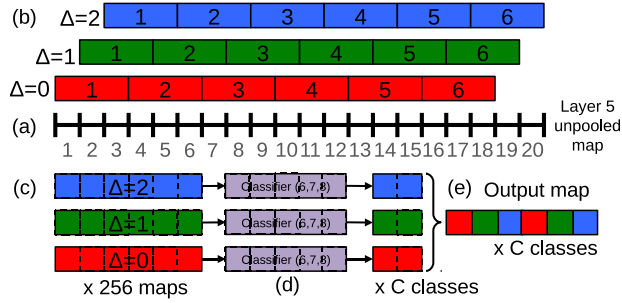}
\end{center}
\vspace*{-0.3cm}
\caption{1D illustration (to scale) of output map computation for classification, using
  $y$-dimension from scale 2 as an example (see
  \tab{multiscale}). (a): 20 pixel unpooled layer 5 feature map. (b):
  max pooling over non-overlapping 3 pixel groups, using offsets of
  $\Delta=\{0,1,2\}$ pixels (red, green, blue respectively). (c): The
  resulting 6 pixel pooled maps, for different $\Delta$. (d): 5 pixel
  classifier (layers 6,7) is applied in sliding window fashion to pooled maps,
  yielding 2 pixel by $C$ maps for each $\Delta$. (e): reshaped into 6
  pixel by $C$ output maps.}
\label{fig:classify}
%\vspace*{-0.6cm}
\end{figure}

We now explain in detail how the resolution augmentation is performed.
We use 6 scales of input which result in unpooled layer 5 maps of
varying resolution (see \tab{multiscale} for details). These are then
pooled and presented to the classifier using the following procedure,
illustrated in \fig{classify}:
\begin{enumerate}[(a)]
\item For a single image, at a given scale, we start with the unpooled layer 5
  feature maps.  
\vspace{-2mm}
\item Each of unpooled maps undergoes a 3x3 max pooling operation
(non-overlapping regions), repeated 3x3 times for $(\Delta_x,\Delta_y)$
pixel offsets of $\{0,1,2\}$.
\vspace{-2mm}
\item This produces a set of pooled feature maps, replicated (3x3)
  times for different $(\Delta_x,\Delta_y)$ combinations.
\vspace{-2mm}
\item  The classifier (layers 6,7,8) has a fixed input size of 5x5 and
produces a $C$-dimensional output vector for each location within the
pooled maps. The classifier is applied in sliding-window fashion to the pooled
maps, yielding $C$-dimensional output maps (for a given
$(\Delta_x,\Delta_y)$ combination).
\vspace{-2mm}
\item The output maps for different $(\Delta_x,\Delta_y)$ combinations
  are reshaped into a single 3D output map (two spatial dimensions x
  $C$ classes).
\end{enumerate}

% The maps at
% a given scale then undergo a 3x3 max pooling operation
% (non-overlapping regions), repeated 3x3
% times for $(Delta_x,Delta_y)$ pixel offsets of $\{0,1,2\}$. For each
% $(Delta_x,Delta_y)$ combination, have a set of pooled maps which are
% passed to the classifier. The classifier has a fixed input size of 5x5 and
% produces a $C$-dimensional output vector for location within the
% pooled maps. As the feature maps are larger than 5x5, the 
% classifier is applied in a sliding window fashion, yeilding a 3D output from the
% network at each scale.   layers which
% have a fixed , we , with the resolutions carefully selected to give
% integer sizes for the top convolutional feature maps (the 5th layer of
% our model). These range from 5x5 for a 245x245 input image (same size
% as \cite{Kriz12}) to 11x14 for a 461x569 pixel image (see
% \tab{multiscale} for details). 

% The classifier layers (6th, 7th and output) take 5x5 regions from the
% top convolutional feature maps and project them to a $C$ dimensional
% output vector. As the feature maps are larger than 5x5, the 
% classifier is applied in a sliding window fashion, yeilding a 3D output from the
% network at each scale. For example scale 2 has layer 5 feature maps of size 10x13,
% which gives an output of size 6x9x$C$\footnote{As
%   (10-5+1)x(13-5+1)=6x9.}.

These operations can be viewed as shifting the classifier's viewing window by 1
pixel through pooling layers without subsampling and using skip-kernels in the
following layer (where values in the neighborhood are non-adjacent).  Or
equivalently, as applying the final pooling layer and fully-connected stack at
every possible offset, and assembling the results by interleaving the outputs.

The procedure above is repeated for the horizontally flipped version
of each image. We then produce the final classification by (i) taking the
spatial max for each class, at each scale and flip; (ii) averaging the
resulting $C$-dimensional vectors from different scales and flips and (iii)
taking the top-1 or top-5 elements  (depending on the evaluation
criterion) from the mean class vector. 

At an intuitive level, the
two halves of the network --- i.e.~feature extraction layers (1-5)
and classifier layers (6-output) --- are used in opposite ways. In the
feature extraction portion, the filters are convolved across the
entire image in one pass. From a computational perspective, this is far
more efficient than sliding a fixed-size feature extractor over the
image and then aggregating the results from different
locations\footnote{Our network with 6 scales takes around 2 secs on a
  K20x GPU to process one image}. However, these principles are reversed for the classifier
portion of the network. Here, we want to hunt for a fixed-size
representation in the layer 5 feature maps across different positions
and scales. Thus the classifier has a fixed-size 5x5 input and is
exhaustively applied to the layer 5 maps. The exhaustive
pooling scheme (with single pixel shifts $(\Delta_x,\Delta_y)$)
ensures that we can obtain fine alignment between the classifier and
the representation of the object in the feature map. 
%is different from doing 3x3 pooling without subsampling stride 1, as
%in the this case the fixed-size classifier would see a
%portion of the input image a third the size.

% {\bf Computation issues:} When computing the layer 1-5 features at each scale,
% the filters are convolved across the entire image in one pass, rather than
% using a sliding window of fixed scale and then combining
% features. This is significantly more efficient and allows us compute
% the outputs for each image in 2 secs on an nVidia K20x GPU. 

\subsection{Results}
\label{sec:class_results}
In \tab{clsval}, we experiment with different approaches,
and compare them to the single network model of Krizhevsky
\etal \cite{Kriz12} for reference. The approach described
above, with 6 scales, achieves a top-5 error rate of 13.6\%.
As might be expected, using fewer scales hurts performance: the
single-scale model is worse with 16.97\% top-5 error. The fine stride technique
illustrated in \fig{classify}
brings a relatively small improvement in the single scale regime, but is also
of importance for the multi-scale gains shown here.

%No results demonstrate this claim at the time of writing but this will
%be addressed in the final paper.

\begin{table}[h!]
\small
%\scriptsize
%\vspace*{-2mm}
\begin{center}
\begin{tabular}{|l||c|c|}
  \hline  
           & Top-1    & Top-5  \\
  Approach & error \% & error \%\\ \hline
  \hline
  Krizhevsky \etal \cite{Kriz12}                             & 40.7 & 18.2 \\ \hline \hline
  OverFeat - 1 {\em fast} model, scale 1, coarse stride            & 39.28 & 17.12 \\ \hline 
  OverFeat - 1 {\em fast} model, scale 1, fine stride              & 39.01 & 16.97   \\ \hline 
  OverFeat - 1 {\em fast} model, 4 scales (1,2,4,6), fine stride   & 38.57 & 16.39 \\ \hline
  OverFeat - 1 {\em fast} model, 6 scales (1-6), fine stride       & 38.12 & 16.27 \\ \hline
  OverFeat - 1 {\em accurate} model, 4 corners + center + flip          & 35.60 & 14.71 \\ \hline
  OverFeat - 1 {\em accurate} model, 4 scales, fine stride              & 35.74 & 14.18 \\ \hline
  OverFeat - 7 {\em fast} models, 4 scales, fine stride            & 35.10 & 13.86 \\ \hline
  OverFeat - 7 {\em accurate} models, 4 scales, fine stride             & 33.96 & 13.24 \\ \hline
\end{tabular}
\vspace*{2mm}
\caption{\textbf{Classification experiments on validation set.}
  Fine/coarse stride refers
  to the number of $\Delta$ values used when applying the
  classifier. Fine: $\Delta={0,1,2}$; coarse: $\Delta=0$. }
\label{tab:clsval}
%\vspace*{-4mm}
\end{center}
\end{table}

\begin{figure}[t!]
\vspace{-3mm}
\begin{center}
\includegraphics[width=1\linewidth]{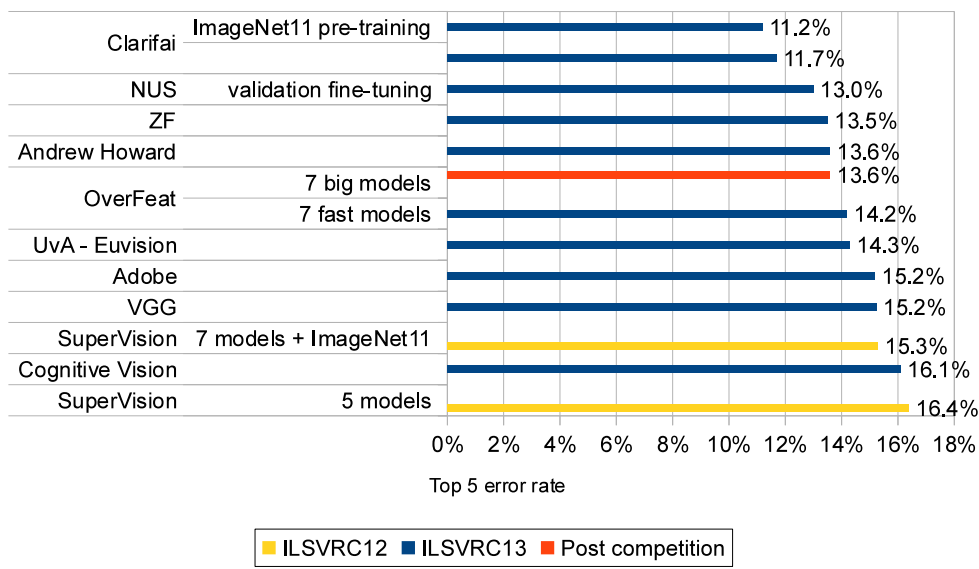}
\end{center}
\vspace*{-0.3cm}
\caption{\textbf{Test set classification results.}
During the competition, OverFeat yielded 14.2\% top 5 error rate using an average
 of 7 fast models.
In post-competition work, OverFeat ranks fifth with 13.6\% error
using bigger models (more features and more layers).}
\label{fig:cls}
%\vspace*{-0.6cm}
\end{figure}

We report the test set results of the 2013 competition in \fig{cls}
where our model (OverFeat) obtained 14.2\% accuracy by voting of 7 ConvNets
(each trained with different initializations) and ranked 5th out of 18 teams.
The best accuracy using only ILSVRC13 data was
11.7\%. Pre-training with extra data from the ImageNet Fall11 dataset improved
this number to 11.2\%. In post-competition work, we improve the OverFeat results down
to 13.6\% error by using bigger models (more features and more layers).
Due to time constraints, these bigger models are not fully trained, more improvements
are expected to appear in time.

%% \begin{table}[h!]
%% \small
%% %\scriptsize
%% %\vspace*{-2mm}
%% \begin{center}
%% \begin{tabular}{|c|c|c|c|}
%%   \hline  
%%   Team & Details & Top-5 error \% & Extra data\\ \hline
%%   \hline
%%   Krizhevsky\etal \cite{Kriz12} & 5 models & 16.4 & \\ \hline
%%   CognitiveVision & & 16.1 & \\ \hline
%%   OverFeat & 1 model & 15.7 & \\ \hline
%%   Krizhevsky \etal \cite{Kriz12} & 7 models & 15.3 & ImageNet Fall11 \\ \hline
%%   VGG & & 15.24 & \\ \hline
%%   Adobe & & 15.19 & \\ \hline
%%   UvA-Euvision & & 14.3 & \\ \hline
%%   OverFeat & 7 models & 14.2 &  \\ \hline
%%   Andrew Howard & & 13.6 &  \\ \hline
%%   ZF & & 13.5 & \\ \hline
%%   NUS & & 13.0 & \\ \hline
%%   Clarifai & & 11.7 & \\ \hline
%%   Clarifai & & 11.2 & ImageNet Fall11 \\ \hline
%% \end{tabular}
%% \vspace*{2mm}
%% \caption{\textbf{ILSVRC13 test set Classification results.}}
%% \label{tab:classify13}
%% %\vspace*{-4mm}
%% \end{center}
%% \end{table}

\subsection{ConvNets and Sliding Window Efficiency}
\label{sec:convnet}

\begin{figure}[t!]
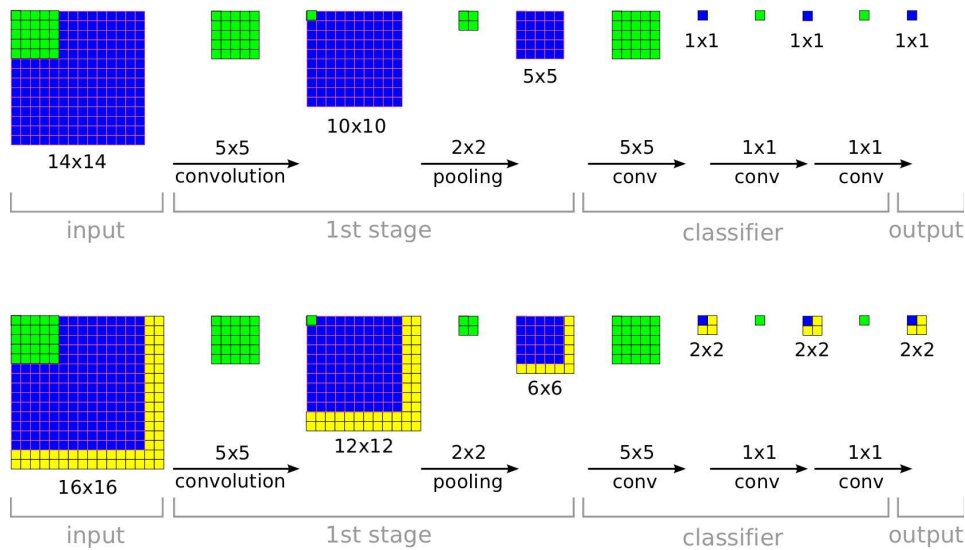

\vspace{-3mm}
\begin{center}
\includegraphics[width=5in]{\dirnet{conv_det0}} \\
%\vspace{2mm}
\includegraphics[width=5in]{\dirnet{conv_det1}} \\
%% %\vspace{2mm}
%% %\includegraphics[width=5in]{\dirnet{conv_det2}} \\
\end{center}
\vspace*{-0.3cm}
\caption{\textbf{The efficiency of ConvNets for detection.} During training,
a ConvNet produces only a single spatial output (top).
But when applied at test time over a larger image, it produces a spatial output
map, e.g. 2x2 (bottom). Since all layers are applied convolutionally,
the extra computation required for the larger image is limited to the yellow regions.
This diagram omits the feature dimension for simplicity.
}
\label{fig:convnet}
%\vspace*{-0.6cm}
\end{figure}

In contrast to many sliding-window approaches that compute an entire pipeline
for each window of the input one at a time, ConvNets are inherently efficient
when applied in a sliding fashion because they
naturally share computations common to overlapping regions.
When applying our network to larger images at test time, we simply
apply each convolution over the extent of the full image.  This extends the output
of each layer to cover the new image size, eventually producing a map of output
class predictions, with one spatial location for each ``window'' (field of view)
of input.  This is diagrammed in \fig{convnet}.  Convolutions are applied bottom-up,
so that the computations common to neighboring windows need only be done once.

Note that the last layers of our architecture are fully connected linear layers.
At test time, these layers are effectively replaced by convolution operations
with kernels of 1x1 spatial extent. The entire ConvNet is then simply a sequence of
convolutions, max-pooling and thresholding operations exclusively.

%%%%%%%%%%%%%%%%%%%%%%%%%%%%%%%%%%%%%%%%%%%%%%%%%%%%%%%%%%%%%%%%%%%%%%%%%%%%%%%

\section{Localization}

Starting from our classification-trained network, we replace the classifier layers
by a regression network and train it to predict object bounding boxes at each
spatial location and scale. We then combine the regression predictions
together, along with the classification results at each location, as we now describe. 

\subsection{Generating Predictions}

To generate object bounding box predictions, we simultaneously
run the classifier and regressor networks across all locations and scales.
Since these share the same feature extraction layers, only the final regression
layers need to be recomputed after computing the classification network.
The output of the final softmax layer for a class $c$ at each
location provides a score of confidence that an object of class $c$
is present (though not necessarily fully contained) in the corresponding field
of view. Thus we can assign a confidence to each bounding box.

\begin{figure}[ht!]
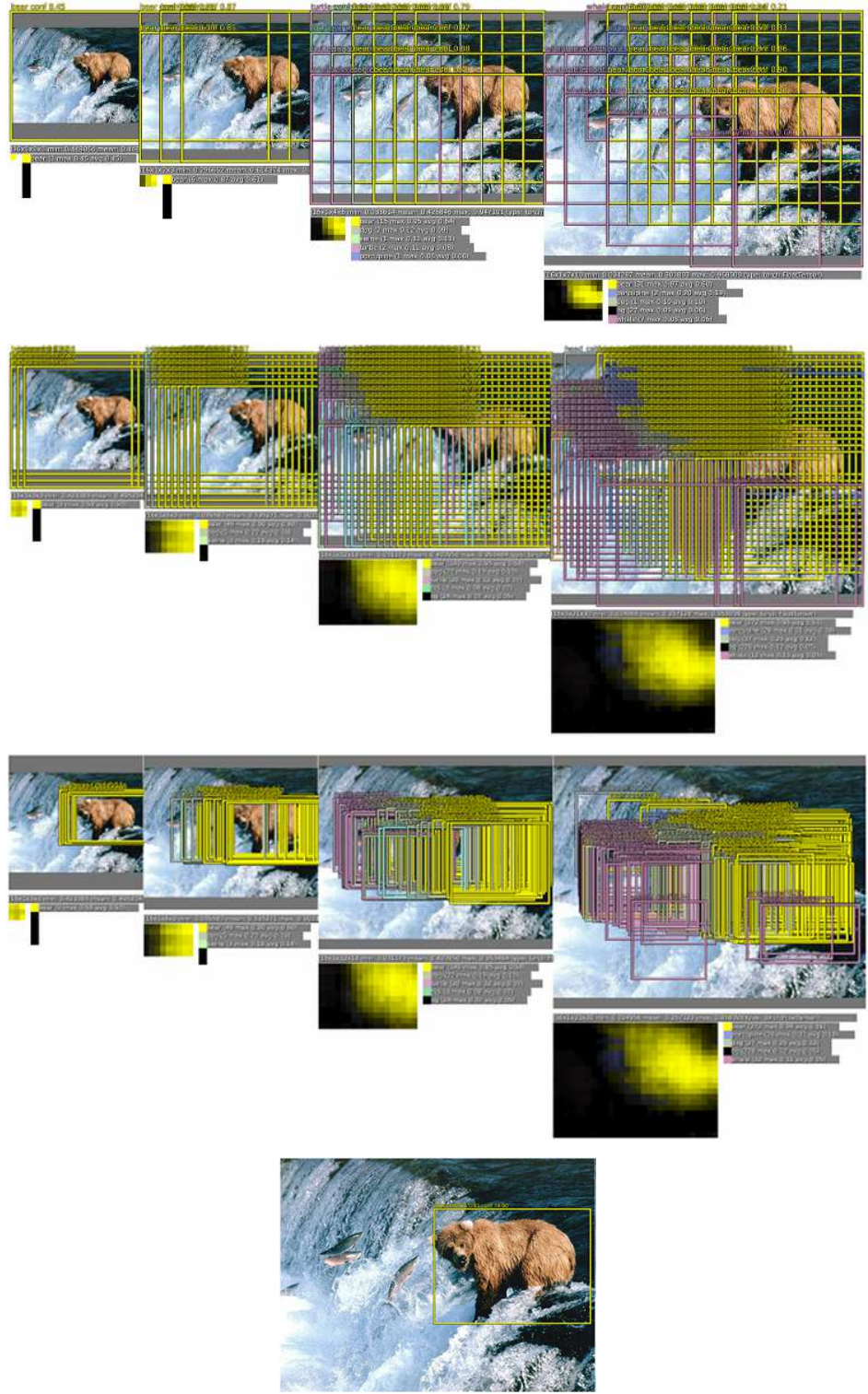

\vspace{-3mm}
\begin{center}
\includegraphics[width=4.9in]{\dirdet{ms_nofine_grid}} \\
\vspace{2mm}
\includegraphics[width=4.9in]{\dirdet{ms_fine_grid}} \\
\vspace{2mm}
\includegraphics[width=4.9in]{\dirdet{ms_fine}} \\
\vspace{2mm}
\includegraphics[width=1.8in]{\dirdet{vote_fine}} \\
\end{center}
\vspace*{-0.3cm}
\caption{\textbf{Localization/Detection pipeline.}
  The raw classifier/detector outputs a class and a confidence for each location
  (1st diagram). The resolution of these predictions can be increased using the
  method described in section~\ref{sec:ms_class} (2nd diagram).
  The regression then predicts the location scale of the object with respect to
  each window (3rd diagram). These bounding boxes are then merge and accumulated
  to a small number of objects (4th diagram).
}
\label{fig:ms}
%\vspace*{-0.6cm}
\end{figure}

\begin{figure}[t!]
\vspace{-3mm}
\begin{center}
\includegraphics[width=5.3in]{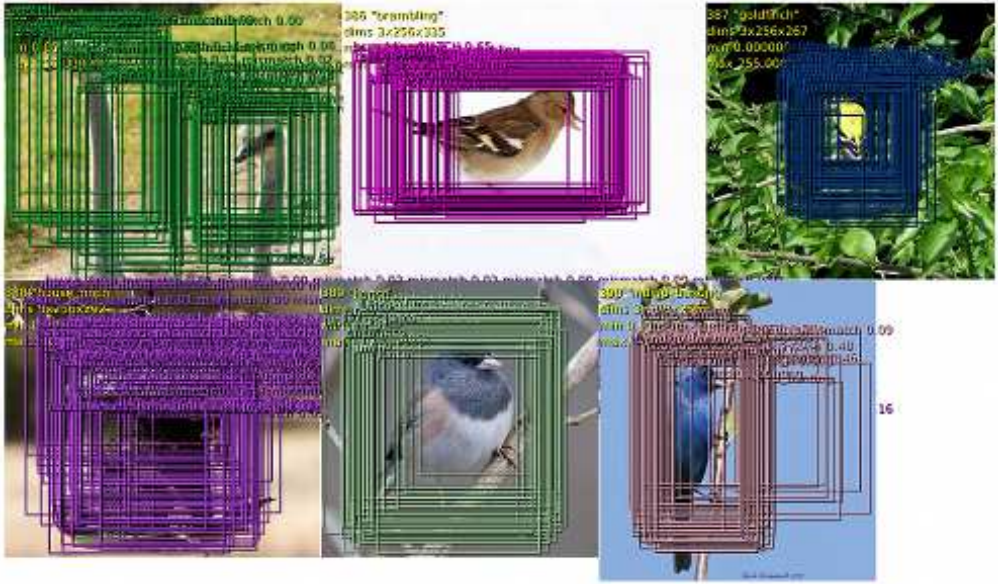} \\
\end{center}
\vspace*{-0.3cm}
\caption{\textbf{Examples of bounding boxes produced by the regression network},
before being combined into final predictions.
The examples shown here are at a single scale. Predictions may be more optimal
at other scales depending on the objects. Here, most of the bounding boxes which
are initially organized as a grid, converge to a single location and scale.
This indicates that the network is very confident in the location of the object,
as opposed to being spread out randomly. The top left image shows that it can also correctly
identify multiple location if several objects are present. The various aspect ratios
of the predicted bounding boxes shows that the network is able to cope with various
object poses.}
\label{fig:examples2}
%\vspace*{-0.6cm}
\end{figure}

\subsection{Regressor Training}

\begin{figure}[t!]
\vspace{-3mm}
\begin{center}
\includegraphics[width=3.3in]{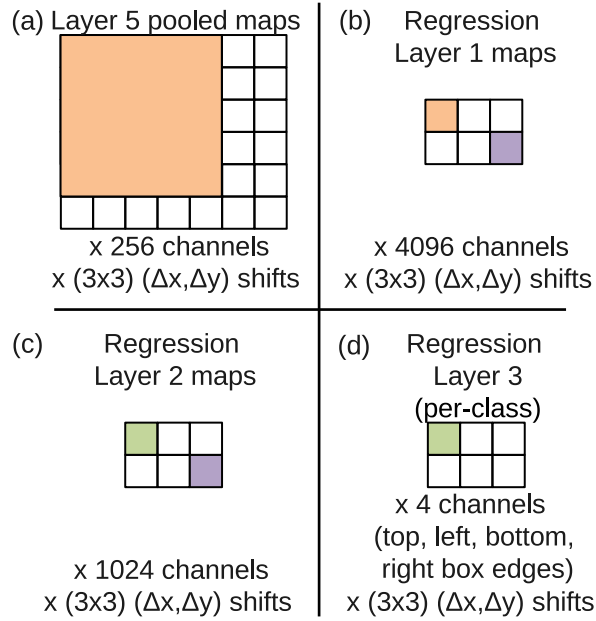}
\end{center}
\vspace*{-0.3cm}
\caption{Application of the regression network to layer 5 features, at
scale 2, for example. (a) The input to the regressor at this scale are 6x7 pixels
spatially by 256 channels for each of
the (3x3) $\Delta_x,\Delta_y$ shifts. (b) Each unit in the 1st layer
of the regression net is connected to a 5x5 spatial neighborhood in the
layer 5 maps, as well as all 256 channels. Shifting the 5x5
neighborhood around results in a map of 2x3 spatial extent, for each
of the 4096 channels in the layer, and for each of
the (3x3) $\Delta_x,\Delta_y$ shifts. (c) The 2nd regression layer has
1024 units and is fully connected (i.e. the purple element only
connects to the purple element in (b), across all 4096 channels). (d)
The output of the regression network is a 4-vector (specifying the
edges of the bounding box) for each location in the 2x3 map, and for
each of the (3x3) $\Delta_x,\Delta_y$ shifts.}
\label{fig:regress}
%\vspace*{-0.6cm}
\end{figure}

The regression network takes as input the pooled feature maps from
layer 5. It has 2 fully-connected hidden layers of size 4096 and 1024
channels, respectively. The final output layer
has 4 units which specify the coordinates for the bounding box
edges.  As with classification, there are (3x3) copies
throughout, resulting from the $\Delta_x,\Delta_y$ shifts. The
architecture is shown in \fig{regress}.

We fix the feature extraction layers (1-5) from the classification network and
train the regression network using an $\ell_2$ loss between the predicted and
true bounding box for each example.  The final regressor layer is
class-specific, having 1000 different versions, one for each class.  We train
this network using the same set of scales as described in \secc{classify}.  We
compare the prediction of the regressor net at each spatial location with the
ground-truth bounding box, shifted into the frame of reference of the
regressor's translation offset within the convolution (see \fig{regress}).
However, we do not train the regressor on bounding boxes with less than 50\%
overlap with the input field of view:
since the object is mostly outside of these locations,
it will be better handled by regression windows that do contain the object.

Training the regressors in a multi-scale manner is important for the
across-scale prediction combination. Training on a single scale will perform well
on that scale and still perform reasonably on other scales.
However training multi-scale will make predictions match correctly across scales and
exponentially increase the confidence of the merged predictions.
In turn, this allows us to perform well with a few scales only, rather than many scales
as is typically the case in detection. The typical ratio from one scale to another
in pedestrian detection~\cite{sermanet-cvpr-13} is about 1.05 to 1.1,
here however we use a large ratio of
approximately 1.4 (this number differs for each scale since dimensions are adjusted
to fit exactly the stride of our network) which allows us to run our system faster.

\subsection{Combining Predictions}

We combine the individual predictions (see \fig{examples2})
via a greedy merge strategy applied to the
regressor bounding boxes, using the following algorithm.

\begin{enumerate}[(a)]
\item  Assign to $C_s$ the set of classes in the top $k$ for each scale $s \in 1\dots 6$, found by taking the maximum detection class outputs across spatial locations for that scale.
\item  Assign to $B_s$ the set of bounding boxes predicted by the regressor network for each class in $C_s$, across all spatial locations at scale $s$.
\item  Assign $B \leftarrow \bigcup_s B_s$
\item  Repeat merging until done:
\item  \quad $(b_1^*, b_2^*) = {\rm argmin}_{b_1\ne b_2 \in B} {\tt match\_score(b_1, b_2)}$
\item  \quad If ${\tt match\_score}(b_1^*, b_2^*) > t$ , stop.
\item  \quad Otherwise, set $B \leftarrow B \backslash \{b_1^*, b_2^*\} \cup {\tt box\_merge}(b_1^*, b_2^*)$
\end{enumerate}

In the above, we compute {\tt match\_score} using the sum of the distance
between centers of the two bounding boxes and the intersection area of the boxes.
{\tt box\_merge} compute the average of the bounding boxes' coordinates.

The final prediction is given by taking the merged bounding boxes with maximum
class scores.  This is computed by cumulatively adding the detection class
outputs associated with the input windows from which each bounding box was
predicted. See \fig{ms} for an example of bounding boxes merged into a single
high-confidence bounding box. In that example, some {\em turtle} and {\em whale}
bounding boxes appear in the intermediate multi-scale steps, but disappear
in the final detection image. Not only do these bounding boxes have low classification
confidence (at most 0.11 and 0.12 respectively), their collection is not as coherent
as the {\em bear} bounding boxes to get a significant confidence boost.
The {\em bear} boxes have a strong confidence (approximately 0.5 on average 
per scale) and high matching scores. Hence after merging, many {\em bear} bounding boxes
are fused into a single very high confidence box, while false positives disappear
below the detection threshold due their lack of bounding box coherence and confidence.
This analysis suggest that our approach is naturally more robust to false positives
coming from the pure-classification model than traditional non-maximum suppression,
by rewarding bounding box coherence.

\subsection{Experiments}

\begin{figure}[t!]
\vspace{-3mm}
\begin{center}
\includegraphics[width=1\linewidth]{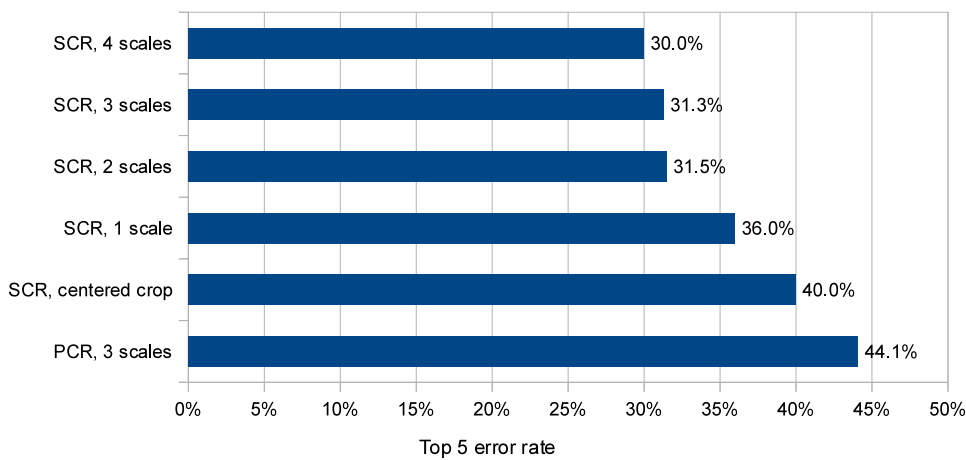}
\end{center}
\vspace*{-0.3cm}
\caption{\textbf{Localization experiments on ILSVRC12 validation set.}
We experiment with different number of scales and with the use of single-class
regression (SCR) or per-class regression (PCR).}
\label{fig:localization_val}
%\vspace*{-0.6cm}
\end{figure}

\begin{figure}[t!]
\vspace{-3mm}
\begin{center}
\includegraphics[width=1\linewidth]{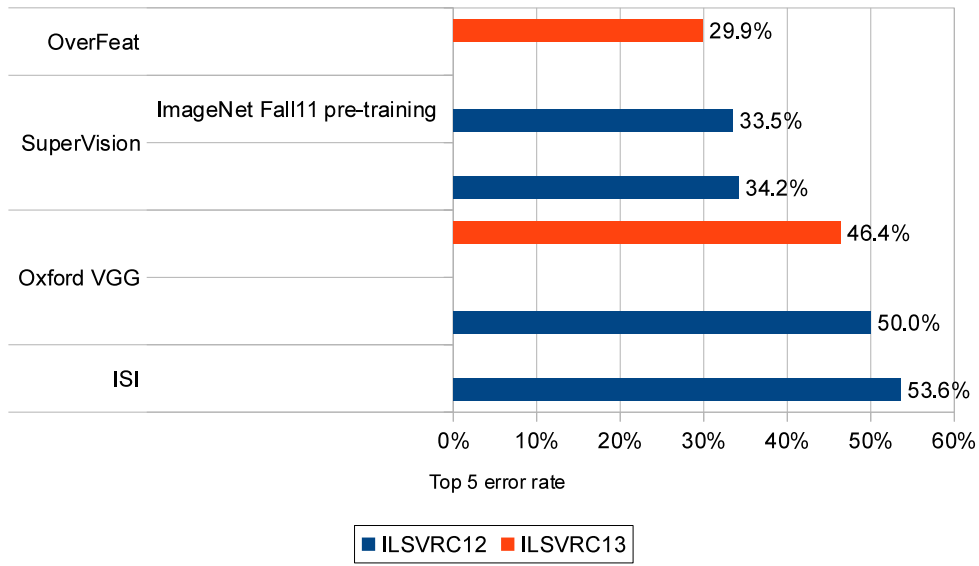}
\end{center}
\vspace*{-0.3cm}
\caption{\textbf{ILSVRC12 and ILSVRC13 competitions results (test set).}
Our entry is the winner of the ILSVRC13 localization competition with 29.9\% error (top 5).
Note that training and testing data is the same for both years.
The OverFeat entry uses 4 scales and a single-class regression approach.}
\label{fig:localization_test}
%\vspace*{-0.6cm}
\end{figure}

We apply our network to the Imagenet 2012 validation set using the
localization criterion specified for the competition. The results for this are shown in
\fig{localization_val}. \fig{localization_test} shows the results of the 2012 and 2013
localization competitions (the train and test data are the same for both of these years).
Our method is the winner of the 2013 competition with 29.9\% error.

Our multiscale and multi-view approach was critical to obtaining good
performance, as can be seen in \fig{localization_val}:  Using only a single centered
crop, our regressor network achieves an error rate of 40\%.  By combining
regressor predictions from all spatial locations at two scales, we achieve a
vastly better error rate of 31.5\%. Adding a third and fourth scale further improves
performance to 30.0\% error.

Using a different top layer for each class in the regressor network for each
class (Per-Class Regressor (PCR) in \fig{localization_val}) surprisingly did not
outperform using only a single network shared among all classes (44.1\% vs.
31.3\%).  This may be because there are relatively few examples per class
annotated with bounding boxes in the training set, while the network has 1000
times more top-layer parameters, resulting in insufficient training.  It is
possible this approach may be improved by sharing parameters only among similar
classes (e.g. training one network for all classes of dogs, another for
vehicles, etc.).

%% \begin{table}[h!]
%% \small
%% %\scriptsize
%% %\vspace*{-2mm}
%% \begin{center}
%% \begin{tabular}{|c|c|c|c|c|}
%%   \hline  
%%   Team & Details & Top-5 error \% & Dataset & Extra data\\ \hline
%%   \hline
%%   VGG & & 46.4 & test &\\ \hline 
%%   OverFeat & PCR, 3 scales & 44.1 & validation & \\ \hline
%%   OverFeat & SCR, single centered crop 221x221 & 40.0 & validation & \\ \hline
%%   OverFeat & SCR, 1 scale & 36.0 & validation & \\ \hline
%%   SuperVision & ILSVRC12 & 34.2 & test & \\ \hline
%%   SuperVision & ILSVRC12 & 33.5 & test & Imagenet Fall 2011 \\ \hline
%%   OverFeat & SCR, 2 scales & 31.5 & validation & \\ \hline
%%   OverFeat & SCR, 3 scales & 31.3 & validation & \\ \hline
%%   OverFeat & SCR, 4 scales & 30.0 & validation & \\ \hline
%%   OverFeat & SCR 4 scales & 29.9 & test &\\ \hline
%% \end{tabular}
%% \vspace*{2mm}
%% \caption{Comparison between our localization method and leading entries
%%   in the ImageNet 2012 and 2013 competitions. SCR=Single class
%%   regressor; PCR=per class regressor.}
%% \label{tab:localization}
%% %\vspace*{-4mm}
%% \end{center}
%% \end{table}

\section{Detection}

Detection training is similar to classification training but in a spatial
manner. Multiple location of an image may be trained simultaneously.
Since the model is convolutional, all weights are shared among all locations.
The main difference with the localization task, is the necessity to predict
a background class when no object is present.
Traditionally, negative examples are initially taken at random for training.
Then the most offending negative errors are added to the training set
in bootstrapping passes. Independent bootstrapping passes render training
complicated and risk potential mismatches between the negative examples collection
and training times. Additionally, the size of bootstrapping passes needs to be
tuned to make sure training does not overfit on a small set.
To circumvent all these problems, we perform negative training on the fly, by
selecting a few interesting negative examples per image such as random ones
or most offending ones. This approach is more computationally expensive, but
renders the procedure much simpler. And since the feature extraction is initially trained
with the classification task, the detection fine-tuning is not as long anyway.

%% \begin{table}[h!]
%% \small
%% %\scriptsize
%% %\vspace*{-2mm}
%% \begin{center}
%% \begin{tabular}{|c|c|c|c|}
%%   \hline  
%%   Team & Details & mAP \% & extra data \\ \hline
%%   \hline
%%   UIUC-IFP & & 1.0 & \\ \hline
%%   Delta & & 6.1 & \\ \hline
%%   SYSU\_Vision & & 7.5 & \\ \hline
%%   GPU\_UCLA & & 9.8 & \\ \hline
%%   SYSU\_Vision & & 10.5 &\\ \hline
%%   Toronto A & & 11.5 &\\ \hline
%%   UvA-Euvision & pure detection + validation fine-tuning & 19.2 & \\ \hline
%%   OverFeat & pure detection & 19.4 & ILSVRC13\\ \hline
%%   NEC-MU & valiation fine-tuning & 20.9 & \\ \hline
%%   UvA-Euvision & detection + classification + validation fine-tuning & 22.6 & \\ \hline
%%   OverFeat & detection + classification (post competition) & 24.3 & ILSVRC13\\ \hline
%% \end{tabular}
%% \vspace*{2mm}
%% \caption{\textbf{ILSVRC13 test set Detection results.}
%% During the competition, UvA ranked first with 22.6 mAP. In post completion work,
%% we score higher at 24.3 mAP.}
%% \label{tab:detection}
%% %\vspace*{-4mm}
%% \end{center}
%% \end{table}

\begin{figure}[t!]
\vspace{-3mm}
\begin{center}
\includegraphics[width=1\linewidth]{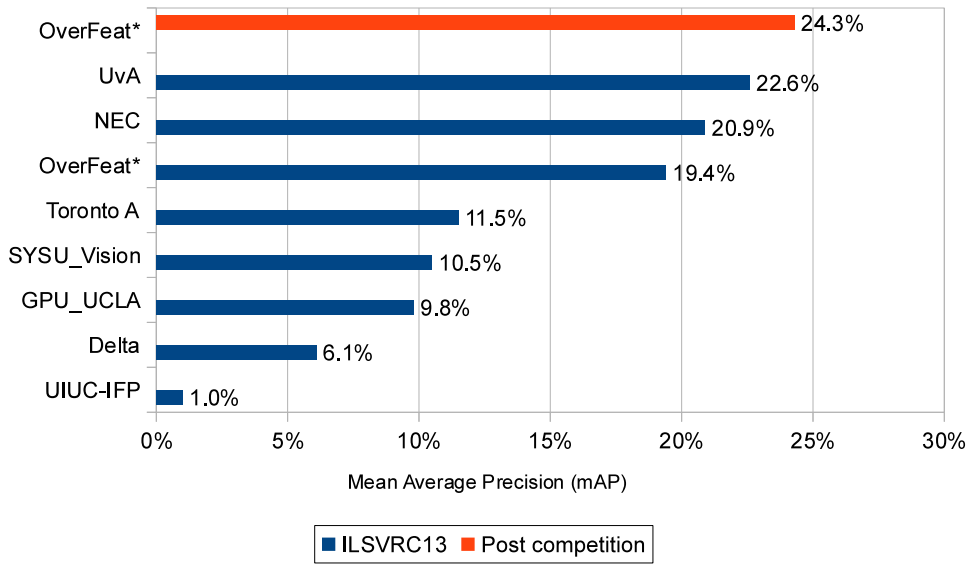}
\end{center}
\vspace*{-0.3cm}
\caption{\textbf{ILSVRC13 test set Detection results.}
During the competition, UvA ranked first with 22.6\% mAP. In post competition work,
we establish a new state of the art with 24.3\% mAP. Systems marked with * were pre-trained with
the ILSVRC12 classification data.}
\label{fig:detection}
%\vspace*{-0.6cm}
\end{figure}

In \fig{detection}, we report the results of the ILSVRC 2013 competition
 where our detection system ranked 3rd with 19.4\% mean average precision (mAP).
We later established a new detection state of the art with 24.3\% mAP.
Note that there is a large gap between
the top 3 methods and other teams (the 4th method yields 11.5\% mAP).
Additionally, our approach is considerably different from the top 2 other systems
which use an initial segmentation step to reduce candidate windows from
approximately 200,000 to 2,000. This technique speeds up inference and substantially
reduces the number of potential false positives.
\cite{UijlingsIJCV2013,carreira2012object}
suggest that detection accuracy drops when using dense sliding window
as opposed to selective search which discards unlikely object locations
hence reducing false positives.
Combined with our method,
we may observe similar improvements as seen here between traditional dense methods
and segmentation based methods.
It should also be noted that we did not fine tune on the detection validation set
as NEC and UvA did. The validation and test set distributions differ significantly
enough from the training set that this alone improves results by approximately 1 point.
The improvement between the two OverFeat results in \fig{detection}
are due to longer training times
and the use of context, i.e. each scale also uses lower resolution scales as input.

\section{Discussion}
\label{sec:discussion}

We have presented a multi-scale, sliding window approach that can be used
for classification, localization and detection.
We applied it to the ILSVRC 2013 datasets, and
it currently ranks 4\textsuperscript{th} in classification, 1\textsuperscript{st}
in localization and 1\textsuperscript{st} in detection.
A second important contribution of our paper
is explaining how ConvNets can be effectively used for detection and
localization tasks. These were never addressed in \cite{Kriz12} and
thus we are the first to explain how this can be done in the context
of ImageNet 2012.  The scheme we propose involves substantial
modifications to networks designed for classification, but clearly
demonstrate that ConvNets are capable of these more challenging
tasks. Our localization approach won the 2013 ILSVRC competition
and significantly outperformed all 2012 and 2013 approaches.
The detection model was among the top performers during the competition,
and ranks first in post-competition results.
We have proposed an integrated pipeline that can perform different tasks
while sharing a common feature extraction base, entirely learned
directly from the pixels.

% have shown ConvNets to be highly effective for classification. In Our
% multi-scale, sliding window approach But it
% has been not been clear how to apply these models to detection and
% localization tasks. In this paper we have explained how this may be
% done efficiently and effectively. The results we obtained on the localization task are
% significantly better than existing non-ConvNet approaches for the
% ImageNet 2012, but are not quite as good as the
% unpublished SuperVision approach.

Our approach might still be improved in several ways.  (i) For localization,
we are not currently back-propping through the whole network; doing so
is likely to improve performance. (ii) We are using $\ell_2$ loss,
rather than directly optimizing the intersection-over-union (IOU) criterion
on which performance is measured. Swapping the loss to this should be
possible since IOU is still differentiable, provided there is some
overlap. (iii) Alternate parameterizations of the bounding box may
help to decorrelate the outputs, which will aid network training. 

{\small
  \bibliographystyle{ieee}
  \bibliography{../bib/bibli,../bib/refs,../bib/bib-bengio,../bib/bib-bibli,../bib/bib-convnets,../bib/bib-deep,../bib/bib-deeplearning,../bib/bib-lecun,../bib/bib-jarrett}

%\bibliography{iclr}

% %\begin{thebibliography}{}
%   \bibliographystyle{ieee}
%   \bibliography{bib-bengio.bbl}
%   \bibliography{bib-convnets.bbl}
%   \bibliography{bib-deeplearning.bbl}
%   \bibliography{bib-jarrett.bbl}
%   \bibliography{refs.bbl}
%   \bibliography{bib-bibli.bbl}
%   \bibliography{bib-deep.bbl}
%   \bibliography{bib-implementation.bbl}
%   \bibliography{bib-lecun.bbl}
% %\end{thebibliography}
}

\newpage
\section*{Appendix:  Additional Model Details}

\begin{table}[h!]
%\small
\scriptsize
%\vspace*{-2mm}
\begin{center}
\begin{tabular}{|l||c|c|c|c|c|c||c|c||c|}
  \hline  
         &     &   &   &   &   &   &   & & Output \\ 
  Layer  &   1 & 2 & 3 & 4 & 5 & 6 & 7 & 8 & 9 \\ \hline
  \hline
  Stage & conv + max & conv + max & conv & conv & conv & conv + max & full & full & full \\ \hline
  \# channels &   96 & 256        & 512  & 512  & 1024 & 1024 & 4096 & 4096  & 1000 \\ \hline
  Filter size &  7x7 & 7x7        & 3x3  & 3x3  & 3x3  & 3x3  & -    & -    & - \\ \hline
  Conv. stride &  2x2 & 1x1       & 1x1  & 1x1  & 1x1  & 1x1  & -    & -    & -  \\ \hline
  Pooling size &  3x3 & 2x2       & -    & -    & -    & 3x3  & -    & -    & -  \\ \hline
  Pooling stride& 3x3 & 2x2       & -    & -    & -    & 3x3  & -    & -    & -  \\ \hline
  Zero-Padding size &  - & - & 1x1x1x1 & 1x1x1x1 & 1x1x1x1 & 1x1x1x1 & - & - & -  \\ \hline
  Spatial input size & 221x221 & 36x36 & 15x15 & 15x15 & 15x15 & 15x15 &  5x5 & 1x1 & 1x1    \\ \hline
 
%  Constrast norm. & - & \checkmark & \checkmark & \xmark
  \hline
\end{tabular}
\vspace*{2mm}
\caption{\textbf{Architecture specifics for {\em accurate} model.}
  It differs from the {\em fast} model mainly in the stride of the first convolution,
  the number of stages and the number of feature maps.}
\label{tab:arch2}
%\vspace*{-4mm}
\end{center}
\end{table}

\begin{table}[h!]
%\small
\scriptsize
%\vspace*{-2mm}
\begin{center}
\begin{tabular}{|c||c|c|}
  \hline  
  model & \# parameters (in millions) & \# connections (in millions) \\  \hline
  Krizhevsky & 60 & - \\ \hline
  {\em fast} & 145 & 2810 \\ \hline
  {\em accurate} & 144 & 5369 \\ \hline
\end{tabular}
\vspace*{2mm}
\caption{\textbf{Number of parameters and connections} for different models.}
\label{tab:connections}
%\vspace*{-4mm}
\end{center}
\end{table}

\begin{table}[h!]
\small
%\scriptsize
%\vspace*{-2mm}
\begin{center}
\begin{tabular}{|c||c|c|c|c||c|}
  \hline  
        & Input & Layer 5 & Layer 5 & Classifier & Classifier \\ 
  Scale & size  & pre-pool & post-pool& map (pre-reshape) & map size \\ \hline \hline
  1     & 245x245 & 17x17 & (5x5)x(3x3)  & (1x1)x(3x3)x$C$ & 3x3x$C$ \\ \hline
  2     & 281x317 & 20x23 & (6x7)x(3x3) & (2x3)x(3x3)x$C$ & 6x9x$C$ \\ \hline
  3     & 317x389 & 23x29 & (7x9)x(3x3) & (3x5)x(3x3)x$C$ & 9x15x$C$ \\ \hline
  4     & 389x461 & 29x35 & (9x11)x(3x3) & (5x7)x(3x3)x$C$ & 15x21x$C$ \\ \hline
  5     & 425x497 & 32x35 & (10x11)x(3x3)  & (6x7)x(3x3)x$C$ & 18x24x$C$ \\ \hline
  6     & 461x569 & 35x44 & (11x14)x(3x3) & (7x10)x(3x3)x$C$ & 21x30x$C$ \\ \hline
\end{tabular}
\vspace*{2mm}
\caption{\textbf{Spatial dimensions of our multi-scale approach}. 6 different
  sizes of input images are used, resulting in layer 5 unpooled
  feature maps of differing spatial resolution (although not indicated
  in the table, all have 256 feature channels). The (3x3) results from
  our dense pooling operation with $(\Delta_x,\Delta_y)=\{0,1,2\}$. See text and
  \fig{classify} for details for how these are converted into output
  maps.}
\label{tab:multiscale}
%\vspace*{-4mm}
\end{center}
\end{table}

\end{document}